\documentclass[11pt,a4paper]{article}

\usepackage[utf8]{inputenc}
\usepackage[T1]{fontenc}
\usepackage{lmodern}
\usepackage[margin=1in]{geometry}
\usepackage{graphicx}
\usepackage{hyperref}
\usepackage{url}
\usepackage{xcolor}
\usepackage{xspace}
\usepackage{etoolbox}
\usepackage{amsmath,amssymb,amsthm}
\usepackage{booktabs}
\usepackage{multirow}
\usepackage{array}
\usepackage{authblk}

\usepackage[authoryear,longnamesfirst]{natbib}

\usepackage{dsfont}
\usepackage{mathtools}
\usepackage{algorithm}
\usepackage{algpseudocode}
\usepackage{longtable}
\usepackage{float}
\usepackage{placeins}

\providecommand{\shorttitle}[1]{}
\providecommand{\shortauthors}[1]{}
\providecommand{\cormark}[1]{\textsuperscript{*}}
\providecommand{\cortext}[2]{}
\makeatletter
\providecommand{\ead}[1]{}
\makeatother
\newenvironment{highlights}{\section*{Highlights}\begin{itemize}}{\end{itemize}}
\newenvironment{keywords}{\vspace{0.5em}\noindent\textbf{Keywords:} }{\par\vspace{0.5em}}
\providecommand{\sep}{,\ }

\def\tsc#1{\csdef{#1}{\textsc{\lowercase{#1}}\xspace}}
\tsc{WGM}
\tsc{QE}

\newcommand{\method}{\textsc{PathBoost}}

\newcommand{\tud}{\textsc{TUDatasets}}

\begin{document}

\title{Path-Based Gradient Boosting for Graph-Level Prediction}

\author[1]{Claudio Meggio\thanks{Corresponding author: \texttt{claudm@math.uio.no}}}
\author[1]{Johan Pensar}
\author[1]{Riccardo De Bin}
\affil[1]{Department of Mathematics, University of Oslo, Oslo, Norway}

\date{}

\begin{abstract}
We propose \method{}, a gradient tree boosting method for graph-level classification and regression that learns discriminative path-based features directly from the input graph structure. Building on a previous work, which was tailored to a specific chemistry application, \method{} introduces three key extensions: (i)~adaptation to binary classification through gradient boosting with a logistic loss, (ii)~incorporation of multiple node and edge attributes into the path feature space via a prefix-based decomposition, and (iii)~automatic anchor node selection based on categorical attribute diversity, eliminating the need for the user to specify the starting point of the considered path features. We compared \method{} to graph neural networks and graph kernel approaches on several benchmark datasets, obtaining better results in half of them, and comparable results in the rest. \method{} shows better performances on graphs with larger average node counts. Overall, the results demonstrate that path-based boosting methods can be competitive with more complex black-box approaches.
\end{abstract}


\begin{highlights}
  \item Gradient boosting framework for graph-level prediction using path features
  \item Prefix decomposition enriches paths with node and edge attributes
  \item Automatic anchor node selection removes need for domain knowledge
  \item An empirical study demonstrates competitiveness compared to existing methods
\end{highlights}


\maketitle

\section{Introduction}\label{sec:introduction}

Graph-level classification is a fundamental problem in machine learning with applications ranging from molecular property prediction and drug discovery to social network analysis and program verification. The challenge lies in learning meaningful representations of entire graphs that capture both local structural patterns and global topological properties while maintaining computational efficiency and interpretability.

Recent years have witnessed the dominance of Graph Neural Networks (GNNs) in graph classification benchmarks. These models leverage message-passing mechanisms~\citep{gilmer2017neural} to learn node embeddings, which are then aggregated to produce graph-level representations. At each layer, nodes aggregate information from their neighbors and update their representations. Graph Convolutional Networks~\citep{kipf2017gcn} 
introduced spectral-inspired convolutional filters on graphs, while Graph Attention Networks~\citep{velickovic2018gat} incorporated attention mechanisms to learn adaptive neighborhood weighting. GraphSAGE~\citep{hamilton2017sage} proposed sampling-based aggregation for inductive learning on large graphs. While GNNs have achieved impressive empirical results, they present several limitations: (i)~they often require substantial amounts of training data to generalize effectively \citep{Ding2024}, (ii)~their learned representations can be difficult to interpret, making it challenging to understand which graph substructures drive predictions \citep{Agarwal2023}, and (iii)~their performance on small datasets with limited samples remains inconsistent \citep{10651061}.

Before the rise of GNNs, graph kernels combined with SVMs were the standard approach for graph classification and regression. These methods explicitly engineer graph features based on domain knowledge, such as counting specific substructures, comparing shortest paths, or measuring graph edit distances. The Weisfeiler--Leman (WL) subtree kernel~\citep{Weisfeiler_1} iteratively refines node labels by hashing neighborhood information, producing subtree-pattern features that approximate the WL isomorphism test. The WL optimal assignment variant~\citep{kriege2016valid} improves the original variant by matching node features via optimal assignment between labeled node hierarchies. Shortest-path kernels~\citep{shortest_path} compare graphs based on the distributions of shortest-path lengths between node pairs, while graphlet kernels~\citep{graphlet} count small induced subgraphs to capture local structural motifs. A systematic comparison by \citet{TUD_paper} demonstrated that graph kernels remain highly competitive on small-scale datasets, often matching or exceeding contemporary GNN performance. This finding motivates the continued exploration of non-neural approaches that can combine the benefits of explicit feature construction with flexible learning algorithms. While interpretable and well-founded theoretically, kernel methods face their own challenges:  (i)~they may fail to capture complex patterns without the right kernel choice, and (ii)~computational costs can become prohibitive for large graphs or datasets \citep{Kriege2020}.

Gradient boosting represents an alternative paradigm that bridges these two approaches: it combines the interpretability of explicit feature construction with the adaptive learning of gradient-based methods. Several approaches have applied boosting to graph or substructure-based classification and regression. ~\citet{KudoAl2004} proposed boosting over labeled subgraphs using a pattern-growth mining algorithm, where subgraph features are first enumerated and then used as weak learners in an AdaBoost framework. ~\citet{SaigoAl2009} extended this to gBoost, which integrates mathematical programming with boosting to select discriminative subgraph patterns. While the former only works for classification tasks, the latter has been implemented also for regression.

A common characteristic of these boosting methods is that they follow a two-stage approach: first enumerating candidate graph patterns, then applying boosting over the enumerated features. In a recent analysis of transition metal compounds, \citet{meggio2024pathboost} developed a boosting approach for regression on molecular graphs that handle graph exploration and model fitting \emph{simultaneously}. At each boosting iteration, the algorithm selects the most informative path and only then expands the feature space by considering one-node extensions of that path. This lazy expansion avoids the combinatorial cost of upfront enumeration and naturally focuses the search on informative regions of the graph. However, the approach of \citet{meggio2024pathboost} was developed to analyse a specific class of graphs, known as transition metal compounds (TMCs), with specific characteristics. Specifically, a TMC contains a metal centre that can be used as a designated anchor node, that is, starting point for the considered path features. Moreover, the original approach focused on the regression setting and did not allow for node or edge attributes. 

In this work we extend the original algorithm to a more general setting as follows: (i)~we enable classification through gradient boosting with logistic loss, (ii)~we incorporate multiple node and edge attributes as part of the path features, and (iii)~we automatically select an anchor node based on node label diversity. These extensions enable our method to compete with state-of-the-art GNNs and graph kernels on graph classification benchmarks while preserving the interpretability advantages of the original framework. An empirical evaluation on \tud{} benchmarks demonstrates that our novel approach achieves competitive or superior performance compared to established competitors across multiple datasets. 

The remainder of this paper is organized as follows. Section~\ref{sec:method} starts by briefly recalling the original algorithm, and then presents the novel extensions, namely the formalization of attribute-enriched path features, automatic anchor node selection, and the adaptation to the classification setting. Section~\ref{sec:experimental-setup} describes the experimental setup, including the datasets selected for the comparison and the competing methods. Section~\ref{sec:results} reports empirical results and comments on the findings. Finally, Section~\ref{sec:discussion} concludes the paper with some discussions on the strengths and weaknesses of the proposed method.

\section{Method}\label{sec:method}

We first summarize the original method by \citep{meggio2024pathboost}, and then present the extensions behind the new method, \method{}, a general supervised learning approach for graph-structured input data.

\subsection{Background}\label{subsec:recap}

\citet{meggio2024pathboost} introduced a boosting-based framework
for regression tasks on graph-structured data, with a particular focus on
molecular graphs representing transition metal compounds. We summarize the
key elements of that framework here, using the same notation, which we extend in
subsequent sections for the classification and attribute-enriched setting
of \method{}.

\paragraph{Graph representation.}
We consider node-labelled undirected graphs defined as triples
$G = (V, E, L)$, where $V = \{v_1, \ldots, v_d\}$ is the set of nodes,
$E \subset V \times V$ is the set of undirected edges, and
$L\colon V \to \mathcal{L}$ is a labeling function that assigns each node a
label from a finite set $\mathcal{L} = \{l^{(1)}, \ldots, l^{(K)}\}$.
In the original application, nodes represent atoms, edges represent bonds,
and labels correspond to atomic symbols.

\paragraph{Paths and labelled paths.}
A path of length $m{-}1$ in a graph is a sequence of distinct nodes
$(v_{j_1}, v_{j_2}, \ldots, v_{j_m})$, where the indices $j_1,\dots, j_m$ form a subset of $1, \dots,d$ such that each consecutive pair is
connected by an edge. The corresponding \emph{labelled path} is the
sequence of node labels
\begin{equation}\label{eq:labelled_path}
  \boldsymbol{l}_u
  = \big(L(v_{j_1}),\, L(v_{j_2}),\, \ldots,\, L(v_{j_m})\big),
\end{equation}
where $u$ indexes the labelled path. We write $M[G_i, \boldsymbol{l}_u]$
for the number of times the labelled path $\boldsymbol{l}_u$ occurs in
graph~$G_i$.

\paragraph{Anchor nodes.}
The original framework assumes a designated anchor node $v^*$ in each
graph, from which all paths originate. In the molecular setting of the
original work, this anchor node is the metal centre of the compound.

\paragraph{Feature mapping and additive model.}
Given a collection of labelled paths
$\mathcal{P} = \{\boldsymbol{l}_u\}_{u=1}^p$, each graph $G_i$ is mapped
to a count vector:
\begin{equation}\label{eq:count_vector}
  h(G_i;\, \mathcal{P})
  = \big(M[G_i, \boldsymbol{l}_1],\, \ldots,\, M[G_i, \boldsymbol{l}_p]\big).
\end{equation}
Based on the count-based representation, the prediction model takes an additive form over paths:

\begin{equation}\label{eq:additive_original}
  F(G_i;\, \beta)
  = \sum_{u=1}^{p}\; \sum_{c=0}^{c_u^{\max}}
    \beta_{u,c} \cdot \mathds{1}\big[ \min (M[G_i, \boldsymbol{l}_u],c_u^{\max}) = c\big],
\end{equation}
where $\beta_{u,c}$ is the coefficient associated with observing path
$\boldsymbol{l}_u$ exactly $c$ times, and $c_u^{\max}$ is the maximum
observed count in the training data.

\paragraph{Learning procedure.}
The key idea behind the learning algorithm is to iteratively expand the path-based feature space as part of a standard gradient boosting procedure~\citep{friedman2001greedy,friedman2002stochastic}, 
using tree stumps as
base learners. At each iteration, the algorithm: (i)~computes the
negative gradient of the loss; (ii)~fits a stump to select the most
informative path; (iii)~updates the model. Crucially, when a path is
selected for the first time,
the algorithm expands the feature matrix by adding all one-node 
extensions of that path found in the training data. Concretely, 
for each graph in the training set, the algorithm scans for 
occurrences of the selected path and records which nodes 
immediately extend it; each distinct extension found across 
the dataset becomes a new candidate path in the feature space 
for subsequent iterations.
This iterative expansion of the feature space allows the algorithm to
discover informative substructures without exhaustive enumeration. The core ideas of
this framework---adaptive path selection, boosting over structured features,
and explainable modelling---form the foundation for \method{}.

\paragraph{Variable importance.}
As a boosting algorithm, the framework of \citet{meggio2024pathboost} has a natural
mechanism for evaluating the importance of each path in the final model.
At each boosting iteration, the reduction in the loss function achieved
by the selected path is recorded. Path importance is then defined by
aggregating these contributions across all iterations in which a given
path was selected. Two complementary variants are
proposed by \citet{meggio2024pathboost}: one sums the absolute loss
reductions attributable to each path over all iterations; the other
measures the \emph{relative} reduction, defined as the difference in
loss improvement between the selected path and the second-best candidate
at each iteration. The relative variant avoids inflating the importance
of paths that could be substituted by correlated alternatives, at the
cost of potentially underestimating the importance of groups of similarly
informative paths. Both measures are rescaled to the range $[0, 100]$
relative to the most important path. 


\subsection{Attribute-Enriched Path Features}\label{subsec:attributes}

A central enhancement in \method{} is its ability to incorporate multiple
node and edge attributes as part of the path-based features. While the original
framework focused on frequency-based representations of labelled paths,
\method{} includes attribute-level information,
enabling richer and more discriminative graph representations.

In many graph classification tasks, 
nodes and edges are annotated with diverse
attributes (e.g., atomic number, bond type, electronegativity). These
attributes often carry predictive signal that complements purely structural information. To leverage this, \method{} constructs a feature space where each feature corresponds not only to the presence of a labelled path, but also to additional attributes associated with that path. \method{} adopts a two-step approach. First,
candidate paths are selected based solely on their frequency, following
the principle of the original approach. Only after a path is selected will its associated attributes be used for fitting. We now formalize this
procedure.

\paragraph{Attribute functions.}
We extend the graph representation to $G = (V, E, L, A_V, A_E)$, where
$A_V\colon V \to \mathbb{R}^{q_V}$ maps each node to a vector of $q_V$
node attributes, and $A_E\colon E \to \mathbb{R}^{q_E}$ maps each edge
to a vector of $q_E$ edge attributes. The labelling function $L$ and
the attribute functions $A_V$, $A_E$ coexist: $L$ assigns each node a
categorical label used for path matching, while $A_V$ and $A_E$ provide
continuous or discrete features used for prediction.

\paragraph{Occurrences of a labelled path.}
A labelled path $\boldsymbol{l}_u = (l_1, \ldots, l_m)$ may occur
multiple times in a graph $G_i$, particularly when anchor nodes are
not unique (see Section~\ref{subsec:anchors}). Each occurrence
corresponds to a different sequence of physical nodes and therefore
carries its own attribute values. We denote the $k$-th occurrence
as an actual sequence of nodes
\begin{equation}\label{eq:occurrence}
  \boldsymbol{l}_u^{(k)}(G_i) = (v_{k,1},\, v_{k,2},\, \ldots,\, v_{k,m}),
  \quad k = 1, \ldots, M[G_i, \boldsymbol{l}_u],
\end{equation}
where each $v_{k,j}$ satisfies $L(v_{k,j}) = l_j$.

\paragraph{Sub-path decomposition.}
For a labelled path $\boldsymbol{l}_u = (l_1, \ldots, l_m)$ of length
$m{-}1$, we define its set of \emph{prefixes} (sub-paths) as
\begin{equation}\label{eq:prefixes}
  \mathrm{Prefix}(\boldsymbol{l}_u)
  = \big\{\boldsymbol{l}_u^{[1]},\; \boldsymbol{l}_u^{[2]},\;
    \ldots,\; \boldsymbol{l}_u^{[m]}\big\},
\end{equation}
where $\boldsymbol{l}_u^{[s]} = (l_1, \ldots, l_s)$ denotes the prefix
consisting of the first $s$ labels. Each prefix is itself a valid
labelled path.

When a path $\boldsymbol{l}_u$ of length $m{-}1$ is selected, the
feature set used for fitting includes information from \emph{all} its
prefixes, not only from the full path. This decomposition is key:
a selected path $(l_1, l_2, l_3)$ generates features from the prefixes
$(l_1)$, $(l_1, l_2)$, and $(l_1, l_2, l_3)$, each contributing its
own count and attributes.

\paragraph{Averaged attributes per prefix.}
Since each prefix $\boldsymbol{l}_u^{[s]}$ may occur multiple times in
graph~$G_i$ (potentially from different regions of the graph), we
aggregate the attributes by averaging over all occurrences. For each
prefix $\boldsymbol{l}_u^{[s]}$ with
$M_s \coloneqq M[G_i, \boldsymbol{l}_u^{[s]}]$ occurrences, the averaged
node attributes of the terminal node are
\begin{equation}\label{eq:avg_attr_node}
  \bar{A}_V\big(G_i,\, \boldsymbol{l}_u^{[s]}\big)
  = \frac{1}{M_s} \sum_{k=1}^{M_s} A_V(v_{k,s}),
\end{equation}
which averages the node attribute vector of the $s$-th (terminal) node
across all occurrences of the prefix in~$G_i$. For prefixes of length
two or more ($s \geq 2$), we additionally define the averaged edge
attributes of the last edge,
\begin{equation}\label{eq:avg_attr_edge}
  \bar{A}_E\big(G_i,\, \boldsymbol{l}_u^{[s]}\big)
  = \frac{1}{M_s} \sum_{k=1}^{M_s} A_E(v_{k,s-1},\, v_{k,s}).
\end{equation}
The count $M[G_i, \boldsymbol{l}_u^{[s]}]$ is not averaged and retains its original meaning as the total number of occurrences.

\paragraph{Extended feature vector.}
We define the per-prefix feature vector $\phi$ and the full extended
feature vector $\Phi$. For each prefix $\boldsymbol{l}_u^{[s]}$ in
graph~$G_i$:
\begin{equation}\label{eq:phi}
  \phi\big[G_i,\, \boldsymbol{l}_u^{[s]}\big] =
  \begin{cases}
    \Big(M[G_i, \boldsymbol{l}_u^{[1]}],\;\;
         \bar{A}_V(G_i, \boldsymbol{l}_u^{[1]})\Big)
      & \text{if } s = 1, \\[6pt]
    \Big(M[G_i, \boldsymbol{l}_u^{[s]}],\;\;
         \bar{A}_V(G_i, \boldsymbol{l}_u^{[s]}),\;\;
         \bar{A}_E(G_i, \boldsymbol{l}_u^{[s]})\Big)
      & \text{if } s \geq 2.
  \end{cases}
\end{equation}
The full feature vector for a selected path $\boldsymbol{l}_u$ of length
$m{-}1$ is obtained by concatenating the features of all its prefixes:
\begin{equation}\label{eq:Phi}
  \Phi[G_i,\, \boldsymbol{l}_u]
  = \Big(\phi\big[G_i, \boldsymbol{l}_u^{[1]}\big],\;\;
         \phi\big[G_i, \boldsymbol{l}_u^{[2]}\big],\;\;
         \ldots,\;\;
         \phi\big[G_i, \boldsymbol{l}_u^{[m]}\big]\Big)
  \;\in\; \mathbb{R}^D,
\end{equation}
where $D = m + q_V + (m-1)(q_V + q_E)$ is the total dimension. 
This follows directly from the structure of $\phi$: the first 
prefix ($s=1$) contributes one count feature and $q_V$ root-node 
attributes, giving $1 + q_V$ features; each of the remaining 
$m-1$ prefixes ($s \geq 2$) contributes one count, $q_V$ 
terminal-node attributes, and $q_E$ last-edge attributes, giving 
$1 + q_V + q_E$ features each. Summing over all $m$ prefixes 
yields $(1 + q_V) + (m-1)(1 + q_V + q_E) = m + q_V + (m-1)(q_V + q_E)$.

\paragraph{Design rationale.}
The separation between path selection and model fitting is deliberate and
serves multiple purposes. The selector (a decision stump on count features)
identifies which path is most informative for improving predictions, while
the base learner then fits on the extended feature vector of the selected
path across all graphs in the training set. This two-stage design preserves
interpretability: the selected path provides a clear structural explanation,
regardless of how the base learner uses the associated attributes. If the
selector were allowed to consider all attributes simultaneously, three
problems would arise: (i)~the computational cost would increase substantially
as the selector would need to evaluate all path-attribute combinations;
(ii)~the risk of overfitting would increase, as the selector could exploit
spurious correlations in attribute values; and (iii)~interpretability would
be compromised, as it would become unclear whether a path was selected for
its structural significance or for incidental attribute values.

After fitting the model on the selected path and its associated features,
the candidate pool is expanded by exploring one-step extensions of the
current path, allowing the algorithm to adaptively discover informative
substructures while keeping the feature space manageable.

\subsection{Automatic Anchor Node Selection}\label{subsec:anchors}

In the original framework, anchor nodes defined the starting points for path exploration. This mechanism allowed users to inject domain-specific knowledge by specifying nodes of particular relevance (such as metal centres in transition metal centre compounds). In cases where there is no obvious anchor node this is actually a limitation of the algorithm since it was not possible to initialize it. In \method{}, this mechanism remains available but is no longer required. The algorithm supports a fully general setting where any node in the graph can serve as an anchor. This modification increases flexibility and removes the need for prior knowledge or manual specification.

When anchor nodes are not specified by the user, \method{} selects them automatically using the following procedure. First, the algorithm identifies categorical node attributes that are shared across all nodes in the dataset. Among these, it selects the attribute with the largest
number of distinct classes. This choice maximizes diversity in anchor selection: more classes lead to richer path exploration, whereas fewer classes (e.g., only two) would allow the algorithm to traverse the dataset structure in very few steps, limiting its ability to capture meaningful graph patterns.

This automatic procedure provides a reasonable default, but we still
recommend the pre-specifation of the anchor nodes when domain knowledge is
available. Expert-guided selection can improve both interpretability and
computational efficiency by focusing the search on relevant regions of
the graph. Furthermore, even within the automatically selected attribute,
some classes may carry more predictive importance than others. Users with
domain expertise can prioritize such classes when defining anchor nodes,
enabling a more targeted search.

In cases where the number of potential anchor nodes becomes prohibitively
large, a simple heuristic can reduce complexity: selecting only the least
frequently occurring node labels across the dataset. By prioritizing rare
nodes, the algorithm focuses on structurally distinctive regions of the
graph, which often carry higher predictive value, while keeping
computational cost manageable.
For the experiments reported in this paper, we use the automatic
selection procedure, ensuring a fair
comparison with competitor methods. Specifically, we use anchor node labels whose number of distinct classes is less than 200 within each dataset.

\subsection{Classification Settings}\label{subsec:classification}

We extended the previous algorithm to handle classification tasks, with the current
implementation focused on binary classification where labels are in
$\{0, 1\}$. While our experiments are limited to this setting, the
framework can be extended to multi-class classification with minimal
modifications. The classification procedure follows the principles of gradient
boosting~\citep{friedman2001greedy,hastie2009elements}. 
We use the logistic loss function, which is standard for binary
classification~\citep{hastie2009elements}:
\begin{equation}\label{eq:cross_entropy}
  \ell(y, F(G)) = -\big[y \log \sigma(F(G)) + (1-y) \log(1 - \sigma(F(G)))\big],
\end{equation}
where $F(G)$ denotes the additive model on the logit scale for a graph $G \in \mathcal{G}$, and $\sigma(\cdot) = 1/(1 + e^{-(\cdot)})$ is the sigmoid function (also known as ``expit function''). $F(G)$ is initialised with a constant term based on the log-odds of the
positive class,
\begin{equation}\label{eq:init}
  F_0 = \log\left(\frac{\bar{y}}{1 - \bar{y}}\right),
\end{equation}
where $\bar{y}$ is the proportion of positive examples in the training
set. Initial predictions are obtained by applying the sigmoid function $\hat{y}_{0,i} = \sigma(F_0)$.

At each boosting iteration $m$, the algorithm computes pseudo-residuals,
which represent the negative gradient of the logistic loss with
respect to the current predictions,
\begin{equation}\label{eq:residuals}
  r_{m,i} = y_i - \hat{y}_{m-1,i}.
\end{equation}
These residuals are passed to the path selector, which identifies the
most informative path $\boldsymbol{l}_{u_m}$ for improving the model.
Once a path is selected, the extended feature vector
$\Phi[G_i, \boldsymbol{l}_{u_m}]$ (as defined in
Section~\ref{subsec:attributes}) is constructed, and a base learner
$h_m$ is fitted to the pseudo-residuals. The model is then updated as
\begin{equation}\label{eq:update}
  F_m(G_i) = F_{m-1}(G_i) + \eta \cdot h_m\big(\Phi[G_i, \boldsymbol{l}_{u_m}]\big),
\end{equation}
where $\eta \in (0,1]$ is a tuning parameter called the learning rate. Predictions are updated by applying
the sigmoid function to the new model output.

\subsection{The Extended Additive Model}\label{subsec:extended_model}

In the original framework, the model is a sum of path-specific terms
based solely on occurrence counts
(Equation~\ref{eq:additive_original}), and each base learner is a stump
that makes a single split on a single count variable. In \method{}, the
base learner at each iteration operates on the extended feature vector
$\Phi[G_i, \boldsymbol{l}_{u_m}]$ defined in~\eqref{eq:Phi}, which
incorporates both counts and averaged attributes from all prefixes of
the selected path. Since this feature vector is multi-dimensional, the
base learner can exploit interactions among the attributes within a
single path.

The model after $m_{\mathrm{stop}}$ boosting iterations takes the form
\begin{equation}\label{eq:additive_extended}
  F(G_i)
  = F_0 + \eta \sum_{m=1}^{m_{\mathrm{stop}}}
    h_m\Big(\Phi\big[G_i,\, \boldsymbol{l}_{u_m}\big]\Big),
\end{equation}
where $F_0$ is the initialization term (for classification, the log-odds $\log\big(\bar{y}/(1-\bar{y})\big)$, see Eq. \eqref{eq:init}), $\eta$ is the learning rate,
$\boldsymbol{l}_{u_m}$ is the path selected at iteration~$m$, and $h_m$
is the base learner fitted to the pseudo-residuals at that iteration.

Since the same path may be selected at multiple iterations, the model
can equivalently be written as a sum over distinct selected paths
$\mathcal{P}_{\mathrm{selected}} \subseteq \mathcal{P}$,
\begin{equation}\label{eq:additive_per_path}
  f(G_i)
  = F_0 + \sum_{u \in \mathcal{P}_{\mathrm{selected}}}
    F_u\!\Big(\Phi\big[G_i,\, \boldsymbol{l}_u\big]\Big),
\end{equation}
where $\mathcal{M}_u = \{m \in \{1,\ldots,m_{\mathrm{stop}}\} :
\boldsymbol{l}_{u_m} = \boldsymbol{l}_u\}$ collects the iterations
at which path $\boldsymbol{l}_u$ was selected, and
\begin{equation}\label{eq:f_u_def}
  f_u\!\Big(\Phi\big[G_i,\, \boldsymbol{l}_u\big]\Big)
  = \eta \sum_{m \in \mathcal{M}_u}
    h_m\!\Big(\Phi\big[G_i,\, \boldsymbol{l}_u\big]\Big)
\end{equation}
aggregates their contributions. This formulation generalizes the
coefficient-based representation $\beta_{u,c}$ in the original
framework (Equation~\ref{eq:additive_original}) to the
attribute-enriched setting: each path contributes through a 
function $f_u$ of its extended feature vector rather than through
scalar coefficients on count indicators.

A key property of this formulation is that explainability is preserved
regardless of the choice of base learner. Because the path selector (a
stump fitted on the count matrix) determines which path is relevant at
each iteration, the explainability of the model is grounded in the
selected paths themselves, not in the internal structure of~$h_m$. This
means that $h_m$ can in principle be any supervised learner---a linear
model, a neural network, or a decision tree---without compromising the
path-level explainability.
In our implementation, we use decision trees as base learners. 

There are two key differences from the original model in Equation~\eqref{eq:additive_original}: First, the input to each base learner is the extended feature vector $\Phi[G_i, \boldsymbol{l}_{u_m}]$ rather than a single count value $M[G_i, \boldsymbol{l}_{u_m}]$; Second, the base learner $h_m$ operates on the full extended feature vector $\Phi[G_i, \boldsymbol{l}_{u_m}]$, allowing it to capture interactions among attributes within a single path. Path  \emph{selection} (see step~(ii) in Algorithm~\ref{alg:epb}) still relies solely on the count matrix; the extended features are used only in the fitting step (see step~(iv) in Algorithm~\ref{alg:epb}), as motivated in Section~\ref{subsec:attributes}.
Although the individual base learners $h_m$ can capture within-path
interactions, the model remains additive \emph{across paths}: at each
iteration, only one path contributes to the update.\\

\subsection{The \method{} Algorithm}

The complete \method{} procedure for binary classification is
summarized in Algorithm~\ref{alg:epb}, where steps~\textbf{(ii)}
and~\textbf{(iv)} correspond respectively to the count-based path
selection and the attribute-based fitting described above.

\begin{algorithm}[H]
\caption{\method{} for Binary Classification}\label{alg:epb}
\begin{algorithmic}[1]

\State \textbf{Input:} Training data
  $\mathcal{T} = \{(G_i, y_i)\}_{i=1}^n$ with $y_i \in \{0,1\}$;
  number of iterations $m_{\mathrm{stop}}$; learning rate $\eta$;
  differentiable loss $\ell(\cdot,\cdot)$.

\Statex
\State \textbf{Initialize:}
\State \quad $\mathcal{P} \gets
  \{l : L(v) = l \text{ for some anchor node } v\}$
  \Comment{Initial paths (anchor labels)}
\State \quad $\boldsymbol{X} \gets \boldsymbol{X}_{\mathcal{T},\mathcal{P}}$
  \Comment{Initial count matrix}
\State \quad $\bar{y} \gets \frac{1}{n}\sum_{i=1}^n y_i$; \quad
  $F_0 \gets \log \big(\bar{y}/(1-\bar{y})\big)$
\State \quad $\hat{y}_{0,i} \gets \sigma(F_0)$ for all $i$
\State \quad $\mathcal{P}_{\mathrm{selected}} \gets \varnothing$

\Statex
\For{$m = 1, \ldots, m_{\mathrm{stop}}$}

  \Statex
  \State \textbf{(i)} Compute pseudo-residuals:
    $r_{m,i} \gets y_i - \hat{y}_{m-1,i}$ for all $i$

  \Statex
  \State \textbf{(ii)} \textit{Path selection:} Fit a decision stump on
    $\{r_{m,i}\}_{i=1}^n$ using the count matrix~$\boldsymbol{X}$.
    Let $\boldsymbol{l}' \in \mathcal{P}$ denote the selected path.

  \Statex
  \State \textbf{(iii)} \textit{Feature construction:} Compute the
    extended feature vector $\Phi[G_i, \boldsymbol{l}']$ for all $i$,
    via the prefix decomposition~\eqref{eq:Phi}.

  \Statex
  \State \textbf{(iv)} \textit{Fit base learner:} Train a base learner
    $h_m$ on $\{(\Phi[G_i, \boldsymbol{l}'],\; r_{m,i})\}_{i=1}^n$.

  \Statex
  \State \textbf{(v)} \textit{Update model:}
    $F_m(G_i) \gets F_{m-1}(G_i) + \eta \cdot
      h_m\big(\Phi[G_i, \boldsymbol{l}']\big)$
    for all $i$.

  \Statex
  \State \textbf{(vi)} \textit{Update predictions:}
    $\hat{y}_{m,i} \gets \sigma(F_m(G_i))$ for all $i$.

  \Statex
  \If{$\boldsymbol{l}' \notin \mathcal{P}_{\mathrm{selected}}$}
    \State \textbf{(vii)} \textit{Expand feature space:} Generate new
      paths by extending $\boldsymbol{l}'$ by one node:
      \[
        \mathcal{P}'
        = \big\{(\boldsymbol{l}', l^{(k)})
          : l^{(k)} \in \mathcal{L}
          \;\wedge\;
          (\boldsymbol{l}', l^{(k)}) \text{ exists in } \mathcal{T}
          \big\}.
      \]
    \State \textbf{(viii)} Update:
      $\mathcal{P} \gets \mathcal{P} \cup \mathcal{P}'$;\;
      $\boldsymbol{X} \gets
        (\boldsymbol{X},\, \boldsymbol{X}_{\mathcal{T},\mathcal{P}'})$;\;
      $\mathcal{P}_{\mathrm{selected}} \gets
        \mathcal{P}_{\mathrm{selected}} \cup \{\boldsymbol{l}'\}$.
  \EndIf

\EndFor

\Statex
\State \textbf{Output:} Final model
  $\hat{F}(G) = F_{m_{\mathrm{stop}}}(G)$.
\State \textbf{Prediction:} Classify $G$ as $1$ if
  $\sigma(\hat{F}(G)) > 0.5$, else $0$.

\end{algorithmic}
\end{algorithm}

\paragraph{Variable importance.} \method{} retains both importance variants without modification. The only difference is that the loss function used for computing reductions is the logistic loss (Equation~\ref{eq:cross_entropy}) rather than the squared-error loss used in the original regression setting, which follows directly from the classification adaptation.

\section{Experimental Setup}\label{sec:experimental-setup}


\subsection{The \tud{} Collection}

We evaluate our algorithm on the \tud{} collection~\citep{TUD_paper,the_story_so_far}, a standardized benchmark specifically created for graph-level learning tasks. Several properties of this collection make it well suited for our evaluation: (i) \tud{} provides a standardized ecosystem including Python-based data loaders, kernel and GNN baseline implementations, and evaluation tools, minimizing confounding factors from non-standardized data preprocessing or divergent competitor implementations; (ii) the collection comprises a large number of distinct datasets, of varying sizes from a wide range of applications, allowing for a robust assessment of an algorithm's generalization capabilities across different domains, including cheminformatics subsets involving small molecules and protein structures (e.g., MUTAG, PROTEINS) and bioinformatics datasets; (iii) the collection offers a diverse set of graph models, including complex graphs that may contain discrete or continuous node labels, edge labels, and edge attributes, which vary significantly across different subsets. For our algorithm, having labelled nodes is fundamental, as it relies on node-level labels to distinguish different paths. The inherent variability in features demands that any high-performing algorithm be robust enough to handle both attribute-rich and purely structural data effectively.

\subsection{Dataset Selection}

We performed our comparison on a selection of the datasets included in \tud{}. For our purpose, indeed, it is fundamental to have at least one categorical label for each node. We define a label to be categorical if fewer than 200 different classes for that label are found across the whole dataset, following the spirit of the original framework 
\citep{meggio2024pathboost} where a similar threshold was used 
to exclude continuous or near-continuous attributes.

Moreover, we only selected datasets in which the response is binary, which anyway constitutes the majority of prediction tasks in \tud{}. We have run our algorithm and 4 competing methods (for more details, see Section \ref{sec:baselines}) on all datasets with these characteristics. 
We decided to focus on the balanced datasets, defined as those where the minority class comprises at least 15\% of the samples, as the results for the unbalanced ones were not particularly informative. The prediction for all methods was dominated by the majority class, as often the case in this situation. The analysis of unbalanced datasets requires ad-hoc procedures \citep[see, e.g., ][]{gong2017rhsboost,wang2021review} that are out of the scope of this work.

The reported results are therefore for the following datasets, for which the specific characteristics are reported in Table~\ref{tab:TUD_overview_balanced}:
\begin{itemize}
    \item \textbf{AIDS}~\citep{riesen2008iam,aids2004}: 2000 molecular
      compounds from the NCI AIDS Antiviral Screen, with nodes as atoms
      and edges as bonds, labeled as active or inactive against HIV.
    \item \textbf{BZR}~\citep{Sutherland2003}: benzodiazepine-related
      molecular graphs with binary activity labels.
    \item \textbf{DHFR}~\citep{Sutherland2003}: dihydrofolate reductase
      inhibitor graphs labeled for activity/inactivity.
    \item \textbf{MUTAG}~\citep{Kriege2012SubgraphMK, Debnath1991}: 188
      mutagenic aromatic and heteroaromatic nitro compounds, classified
      by mutagenicity.
    \item \textbf{PROTEINS\_full}~\citep{Dobson2003, Borgwardt}: protein
      structure graphs where nodes represent secondary structure elements
      and edges indicate sequence or spatial adjacency, labeled by protein
      class (enzyme vs.\ non-enzyme). This is the full version with more
      complete graph properties than the smaller PROTEINS variant.
    \item \textbf{PTC\_FM}~\citep{Kriege2012SubgraphMK, Helma2001}:
      molecular graphs from the Predictive Toxicology Challenge, labeled
      for carcinogenicity in female mice.
    \item \textbf{SYNTHETIC}~\citep{NIPS2013_a2557a7}: artificially
      generated graph instances designed to test learning algorithms on
      controlled graph structures.
    \item \textbf{Tox21\_ARE}~\citep{Tox21Challenge2014} (evaluation,
      testing, and training splits): toxicity prediction datasets from the
      Tox21 challenge focused on the antioxidant response element (ARE)
      signaling pathway. We retain these as separate datasets rather than
      merging them to examine whether performance differences arise from
      sample size variation within the same molecular domain; this question
      is explored more systematically through the subsampling analysis in
      Section~\ref{sec:results}.
    \item \textbf{Tox21\_MMP}~\citep{Tox21Challenge2014} (testing and
      training): similarly, these datasets target the matrix
      metalloproteinase (MMP) endpoint from the Tox21 challenge.
\end{itemize}


\begin{table}[htbp]
\caption{Structural and feature characteristics of the 12 balanced 
datasets used in the primary evaluation. \textbf{Graphs} is the total 
number of input graphs in the dataset. \textbf{AvgNodes} and 
\textbf{AvgEdges} refer to the mean number of nodes and edges per 
graph. \textbf{NodeFeat} and \textbf{EdgeFeat} are the number of 
node and edge features per node and edge respectively. 
\textbf{TotalFeat} is their sum. \textbf{ClassPercentages} shows 
the distribution of binary labels. \textbf{CatAttrClasses} indicates 
the number of distinct values in the categorical attribute used for 
anchor node selection.}\label{tab:TUD_overview_balanced}
\centering
\scalebox{0.80}{
\begin{tabular}{l r r r r r r l r}
\toprule
\textbf{Dataset} & \textbf{Graphs} & \textbf{AvgNodes} & \textbf{AvgEdges} & \textbf{NodeFeat} & \textbf{EdgeFeat} & \textbf{TotalFeat} & \textbf{ClassPercentages} & \textbf{CatAttrClasses} \\
\midrule
AIDS           & 2000 & 15.69  & 16.20  & 5  & 1 & 6  & 0: 20\%, 1: 80\%   & 38 \\
BZR            & 405  & 35.75  & 38.36  & 4  & 0 & 4  & 0: 79\%, 1: 21\%   & 10 \\
DHFR           & 756  & 42.43  & 44.54  & 4  & 0 & 4  & $-$1: 39\%, 1: 61\% & 9  \\
MUTAG          & 188  & 17.93  & 19.79  & 1  & 1 & 2  & $-$1: 34\%, 1: 66\% & 7  \\
PROTEINS\_full & 1113 & 39.06  & 72.82  & 30 & 0 & 30 & 1: 60\%, 2: 40\%   & 3  \\
PTC\_FM        & 349  & 14.11  & 14.48  & 1  & 1 & 2  & $-$1: 59\%, 1: 41\% & 18 \\
SYNTHETIC      & 300  & 100.00 & 196.00 & 2  & 0 & 2  & 0: 50\%, 1: 50\%   & 8  \\
Tox21\_ARE\_eval  & 970  & 17.01 & 17.33 & 1 & 1 & 2 & 0: 84\%, 1: 16\% & 25 \\
Tox21\_ARE\_test  & 234  & 21.99 & 22.91 & 1 & 1 & 2 & 0: 79\%, 1: 21\% & 15 \\
Tox21\_ARE\_train & 5670 & 16.28 & 16.52 & 1 & 1 & 2 & 0: 85\%, 1: 15\% & 46 \\
Tox21\_MMP\_test  & 238  & 21.68 & 22.55 & 1 & 1 & 2 & 0: 84\%, 1: 16\% & 15 \\
Tox21\_MMP\_train & 5418 & 17.49 & 17.83 & 1 & 1 & 2 & 0: 84\%, 1: 16\% & 48 \\
\bottomrule
\end{tabular}
}
\end{table}

\subsection{Competing Methods}\label{sec:baselines}

We compare \method{} against the methods already tested on the \tud{} benchmark by \cite{TUD_paper}. These competitors cover two main families: classical graph kernels and graph neural networks (GNNs). 
Despite the recent focus on GNNs, the \tud{} benchmark study 
demonstrated that classical graph kernels combined with SVMs are still 
highly competitive in graph classification tasks, often matching or 
exceeding contemporary GNN performance on small-scale datasets. Here we consider three graph kernels:
\begin{itemize}
    \item \textbf{Weisfeiler--Lehman (WL)}: An efficient and powerful 
      kernel that approximates graph isomorphism by iteratively refining 
      node labels to capture subtree patterns~\citep{Weisfeiler_1}.
    \item \textbf{Shortest-Path (SP)}: Assesses graph similarity by 
      comparing distributions of shortest path 
      lengths~\citep{shortest_path}.
    \item \textbf{Graphlet (GR)}: Counts small induced subgraphs 
      (graphlets) to encode local motifs~\citep{graphlet}.
\end{itemize}
All kernels are paired with SVM classifiers (LIBSVM or LIBLINEAR) and 
tuned using the protocol recommended by \tud{}.

For the GNN, we use GINE-$\epsilon$ (Graph Isomorphism 
Network Enhanced with Edge Features)~\citep{hu2020pretraining}. \citet{xu2019gin} showed that the discriminative power of 
message-passing GNNs is bounded by the Weisfeiler--Leman graph 
isomorphism test, and designed the Graph Isomorphism Network (GIN) to 
achieve this upper bound. GINE extends GIN by incorporating edge 
features into the aggregation step via a two-layer MLP that maps edge 
features to a fixed dimension, combined with node features via 
summation~\citep{hu2020pretraining}. This makes GINE the recommended architecture in the \tud{} 
benchmark for datasets with edge features, which covers all our 
selected datasets. The overall graph embedding is obtained by applying 
mean pooling to the learned node embeddings, followed by a 2-layer MLP 
classifier with dropout $p=0.5$. Hyperparameters are selected via 1-split validation: number of layers $\{1,2,3,4,5\}$, hidden units 
$\{32,64,128\}$, learning rate $0.01$, up to 200 epochs. Despite its 
strong expressiveness, GINE, like other GNNs, does not 
natively provide the kind of feature-level interpretability that 
path-based methods offer, as its learned representations entangle graph 
substructures within dense vector spaces~\citep{Wu2021,Agarwal2023}.
The specific parameters used by each algorithm in cross-validation are reported in Table S1 of the Supplementary Material.

\subsection{Evaluation}
We follow the unified protocol of \tud{}: 10-fold cross-validation 
repeated 10 times with controlled random seeds, reporting mean and 
standard deviation of accuracy. Since many of the datasets exhibit 
class imbalance, we also report the F1-macro score, defined as the 
unweighted mean of per-class F1 scores,
\begin{equation}
    \text{F1}_{\text{macro}} = \frac{1}{N} \sum_{i=1}^{N} \text{F1}_i, 
\ \ \  \text{where}\ \ \ \
     \text{F1}_i = \frac{2 \cdot \text{Precision}_i \cdot 
    \text{Recall}_i}{\text{Precision}_i + \text{Recall}_i}
\end{equation}
and $N$ is the number of classes (in this paper, always 2). By giving equal weight to each 
class regardless of its frequency, F1-macro provides a more balanced 
assessment than accuracy on imbalanced datasets. Note that for the kernel 
methods it was not possible to compute the F1-macro score due to the provided implementation in \tud{}, and therefore this metric is only reported for \method{} and GNN.

\subsection{Additional Experiment: Regression Setting}\label{subsec:regression}

Although in this work we decided to mainly evaluate the algorithm on its novel ability to handle classification tasks, some of the new features of \method{} also affect the regression settings, in particular the ability to include information related to nodes and edges, and the automatic selection of the anchor node when this is not provided by the user. 

To show that both features work as they are supposed to, we apply \method{} with a classical $L_2$ loss to a single dataset, \texttt{alchemy\_full}, also available in the \tud{} repository, and evaluate the performance in terms of Mean Absolute Error (MAE) and R$^2$ on the first target variable (\texttt{target[0]}, namely the dipole moment). The \texttt{alchemy\_full} dataset consists of 202\,579 molecular graphs with up to 29 nodes each, 12 continuous regression targets, a mean of 10.10 nodes and 10.44 edges per graph, and includes node labels, edge labels, and node attributes (3 attributes per node). We do not provide an anchor node to test the ability of \method{} to select one automatically. Moreover, we apply it both by allowing and not allowing the algorithm to retrieve the node attributes, to evaluate if it take advantage of the extra information available.

\subsection{Computational Resources}

All experiments were executed on a Linux server equipped with two Intel Xeon Gold 6226R CPUs (2.90\,GHz, 16 cores each, 64 threads total) and four NVIDIA A10 GPUs (23\,GB VRAM each), running Red Hat Enterprise Linux 8.10. The software environment is anchored by Python 3.11, PyTorch 2.1, and PyTorch Geometric 2.5, running on the CUDA 12.8 framework. GNN training utilized a single GPU, while \method{} and graph kernel competitors ran on CPU only. As experiments were conducted on shared infrastructure, reported wall-clock times may exhibit variability.

\section{Results}\label{sec:results}




\subsection{Classification}

\begin{table}[H]
\caption{Accuracy comparison on balanced datasets. Results show mean 
accuracy $\pm$ standard deviation over 10-fold cross-validation with 
10 repetitions. Column headers denote: \textbf{PB} = \method{}, 
\textbf{GNN} = GINE, \textbf{WL} = Weisfeiler--Lehman kernel, 
\textbf{GR} = Graphlet kernel, \textbf{SP} = Shortest-Path kernel. 
Best results for each dataset are shown in bold. ``Time limit'' 
indicates methods that exceeded the 20-hour computational budget. 
$^*$ the standard kernel (non-linear) exceeded the time limit; the 
value reported is obtained using a linear kernel instead. The complete 
results with all the linear kernels are reported in 
Table S2 in the Supplementary Material.}
\label{tab:main_results}
\centering
\scalebox{1.}{
\begin{tabular}{l l l l l l}
\toprule
\textbf{Dataset} & \textbf{PB} & \textbf{GNN} & \textbf{WL} & \textbf{GR} & \textbf{SP} \\
\midrule
AIDS              & $\mathbf{99.33 \pm 0.09}$ & $96.66 \pm 0.15$  & $96.42 \pm 0.33$          & $99.04 \pm 0.09$          & $99.02 \pm 0.11$ \\
BZR               & $86.76 \pm 0.55$          & $82.51 \pm 1.37$  & $\mathbf{86.90 \pm 1.20}$ & $82.30 \pm 0.91$          & $82.30 \pm 0.91$ \\
DHFR              & $78.19 \pm 0.76$          & $80.03 \pm 1.08$  & $\mathbf{81.45 \pm 1.22}$ & $74.52 \pm 0.72$          & $74.43 \pm 0.90$ \\
MUTAG             & $\mathbf{89.11 \pm 1.08}$ & $82.83 \pm 2.18$  & $76.35 \pm 2.11$          & $85.32 \pm 0.82$          & $85.32 \pm 0.82$ \\
PROTEINS\_full    & $\mathbf{79.00 \pm 0.51}$ & $71.50 \pm 1.21$  & $73.06 \pm 0.71$          & Time limit                & Time limit \\
PTC\_FM           & $61.80 \pm 0.95$          & $59.55 \pm 1.87$  & $\mathbf{62.84 \pm 1.74}$ & $56.64 \pm 1.72$          & $56.64 \pm 1.72$ \\
SYNTHETIC         & $\mathbf{62.27 \pm 1.67}$ & $48.53 \pm 1.86$  & $45.53 \pm 2.62$          & $45.73 \pm 2.53$          & $45.73 \pm 2.53$ \\
Tox21\_ARE\_eval  & $\mathbf{83.37 \pm 0.18}$ & $81.95 \pm 0.76$  & $81.27 \pm 1.33$          & $82.24 \pm 0.50$          & $82.54 \pm 0.51$ \\
Tox21\_ARE\_test  & $79.40 \pm 0.85$          & $79.21 \pm 1.89$  & $78.06 \pm 2.13$          & $\mathbf{80.63 \pm 1.24}$ & $80.20 \pm 1.26$ \\
Tox21\_ARE\_train & $\mathbf{84.79 \pm 0.06}$ & $83.43 \pm 1.32$  & Time limit                & $84.74 \pm 0.08^*$        & $84.74 \pm 0.08^*$ \\
Tox21\_MMP\_test  & $85.84 \pm 0.89$          & $82.01 \pm 2.88$  & $\mathbf{86.31 \pm 2.11}$ & $82.47 \pm 1.53$          & $82.34 \pm 1.20$ \\
Tox21\_MMP\_train & $85.95 \pm 0.18$          & Time limit        & $\mathbf{91.14 \pm 0.19}$ & $84.96 \pm 0.11^*$        & $84.99 \pm 0.12^*$ \\
\bottomrule
\end{tabular}
}
\end{table}

Table~\ref{tab:main_results} reports the results on the selected 12 
datasets. \method{} demonstrates competitive but variable performance. When comparing against the \emph{single best competitor} for each dataset (whether GNN or kernel-based), \method{} achieves superior accuracy on 6 out of 12 datasets, performs comparably (within 1\% accuracy) on 1 dataset, and under-performs on 5 datasets. In particular, 
\method{} achieves its most substantial improvements on datasets with larger graphs and richer feature sets: \texttt{SYNTHETIC} (100 avg nodes, 196 avg edges), where \method{} obtains 62.27\% accuracy compared to 48.53\% for GNN, representing a $+13.74$ percentage point improvement over the best competitor; \texttt{PROTEINS\_full} (39 avg nodes, 30 node features), where \method{} achieves 79.0\% versus 73.06\% for the best kernel competitor ($+5.94$ points); \texttt{AIDS} (16 avg nodes, 5 node features), where \method{} reaches 99.33\% accuracy, marginally outperforming kernel methods at 99.08\%; and \texttt{MUTAG} (18 avg nodes), where \method{} obtains 89.11\% compared to 85.32\% for the best kernel ($+3.79$ points).

In contrast, \method{} performs slightly worse than the best competitor in \texttt{DHFR}, where \method{} achieves 78.19\% versus 81.45\% for WL kernel ($-3.26$ points), representing our largest deficit on balanced datasets. Finally, in \texttt{BZR} and \texttt{PTC\_FM} the performance of \method{} and its best competitor (in both cases the Weisfeiler–Lehman kernel method) are substantially equal, with less than 1\% difference in accuracy.

These accuracy results are confirmed by comparing \method{} and GNN using the F1-macro score (see Table~\ref{tab:main_results_f1}). It seems, therefore, reasonable to surmise that \method{} tends to perform best on larger, structurally complex graphs, as illustrated by its strongest results on \texttt{SYNTHETIC} (100 avg.\ nodes, the largest among the considered datasets) and \texttt{PROTEINS\_full} (39 avg.\ nodes, 30 node features, the richest representation among the considered datasets). In contrast, performance on smaller graphs such as \texttt{MUTAG} and \texttt{PTC\_FM} seems more mixed, with \method{} being competitive but not systematically dominant. No clear relationship emerges between \method{}'s advantage and other dataset properties such as the number of node features, the number of categorical attribute classes, or the total number of graphs. These observations seem to suggest that graph size and structural complexity are the most relevant indicators of when path-based boosting offers advantages over GNN and kernel approaches, though given the limited number of datasets this should be interpreted as an indicative trend rather than a definitive conclusion. 

A separate consideration should be made for the \texttt{Tox21} family datasets: although the differences in accuracy among the methods are not so high, we still notice contradictory results, with \method{} winning on some splits (\texttt{Tox21\_ARE\_evaluation}, \texttt{Tox21\_ARE\_training}) but losing on others (\texttt{Tox21\_MMP\_train}, $-5.19$ points). In this family, we also notice a relatively low F1-macro score, with small inconsistencies with the analysis of the accuracy.

\begin{figure}[H]
    \centering
    \includegraphics[width=0.7\linewidth]{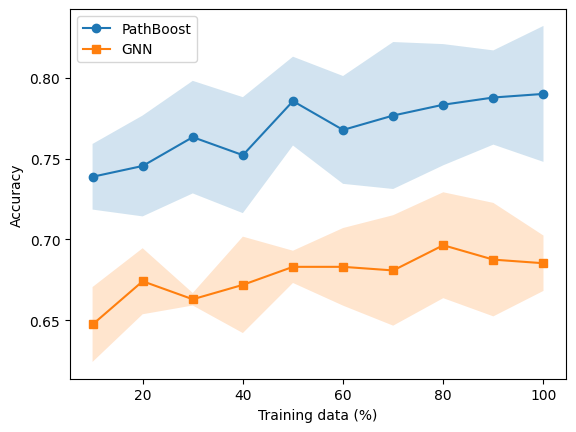}
    \caption{Learning curves on PROTEINS\_full dataset with varying training set sizes. Accuracy of \method{} (upper curve) and GNN competitor (lower curve) as training data increases from 10\% to 100\% of the original sample size. \method{} maintains a consistent performance advantage of approximately 6--8 percentage points across all data regimes, demonstrating that the improvement is systematic rather than a consequence of having access to more training examples.}
    \label{fig:subsample_protein_full}
\end{figure}

To investigate if the contradictory performances can be somehow related to the sample size, we conduct an experiment in which \method{} and GNN are applied to subsamples of increasing size from the PROTEINS\_full. In particular, the sample size was set to vary from 10\% to 100\% of the original sample size, in 10\% increments. Figure~\ref{fig:subsample_protein_full} shows the results: \method{} maintains a consistent performance advantage over GNN across all data regimes, with the accuracy gap remaining approximately constant regardless of the sample size of the training data. This suggests that \method{}'s advantage on PROTEINS\_full does not depend on the sample size. Furthermore, \method{} exhibits similar data efficiency to GNN---neither method shows dramatically steeper or shallower learning curves. The consistent gap indicates a systematic algorithmic difference rather than variance due to sample selection.

\begin{table}[htbp]
\caption{F1 score comparison on balanced datasets. Results show mean $\pm$ standard deviation over 10-fold cross-validation with 10 repetitions. F1-macro averages the per-class F1 scores and provides a class-balanced performance assessment. ``Time limit'' indicates methods that exceeded the 20-hour computational budget. F1-macro comparisons are limited to \method{} and GNN, as the kernel baseline implementations do not allow for metrics beyond accuracy.}
\label{tab:main_results_f1}
\centering
\scalebox{1.0}{
\begin{tabular}{l l l}
\toprule
\textbf{Dataset} & \textbf{PB F1 macro} & \textbf{GNN F1 macro} \\
\midrule
AIDS              & $98.69 \pm 0.12$ & Time limit       \\
BZR               & $70.81 \pm 1.81$ & $69.29 \pm 2.32$ \\
MUTAG             & $83.86 \pm 2.11$ & $80.46 \pm 3.29$ \\
PROTEINS\_full    & $69.14 \pm 1.04$ & Time limit       \\
SYNTHETIC         & $62.00 \pm 19.00$ & $32.00 \pm 0.90$ \\
DHFR              & $75.00 \pm 5.00$  & $78.60 \pm 4.70$ \\      
PTC\_FM           & $51.72 \pm 1.91$ & $53.28 \pm 2.71$ \\
Tox21\_ARE\_eval  & $51.05 \pm 1.99$ & $52.69 \pm 1.67$ \\
Tox21\_ARE\_test  & $54.39 \pm 1.89$ & $59.11 \pm 3.48$ \\
Tox21\_ARE\_train & $50.16 \pm 0.71$ & Time limit       \\
Tox21\_MMP\_test  & $62.14 \pm 3.32$ & $61.14 \pm 4.69$ \\
Tox21\_MMP\_train & $60.83 \pm 0.89$ & Time limit       \\
\bottomrule
\end{tabular}
}
\end{table}



\subsection{Regression Experiment}\label{app:regression}

Table~\ref{tab:regression-ablation} presents the regression results in terms of Mean Absolute Error (MAE) and $R^2$. 
With \method{}, we observe an improvement when incorporating additional node information: the MAE decreases from $1.84\times10^{-2}$ to $1.74\times10^{-2}$, while $R^2$ increases from $0.60$ to $0.64$. On the other hand, this improvement in the prediction comes at the cost of a noticeable increase in the computational time, with the restricted algorithm running roughly $2.5\times$ faster than the complete one, suggesting a favourable accuracy-efficiency trade-off when continuous features are unavailable or costly to compute.

 As a reference, the best performing neural network implemented in the original paper r~\citep{alchemy_full} had a MAE of $2.1\times10^{-2}$, indicating a competitive performance by our approach.
 
  \begin{table}[t]
  \centering
  \caption{Regression performance of \method{} on
  \texttt{alchemy\_full} (target index~0, 10$\times$10-fold CV).
  The complete model (\emph{Complete}) uses all node attributes in addition to the graphical structure, the restricted one (\emph{Restricted}) only the graphical structure.  We report mean $\pm$ standard deviation
  across the 10 repetitions.}
  \label{tab:regression-ablation}
  \begin{tabular}{l c c c}
  \toprule
  \textbf{Variant} & \textbf{MAE} & \textbf{R\textsuperscript{2}}  & \textbf{Time (s)} \\
  \midrule
  Complete             & $0.0174 \pm 0.0000$ & $0.6397 \pm 0.0003$ & $2377.8$ \\
  Restricted         & $0.0184 \pm 0.0000$ & $0.6027 \pm 0.0000$ & $946.7$  \\
  \bottomrule
  \end{tabular}
  \end{table}




\subsection{Computational considerations.}
Systematic timing measurements were not captured owing to the 
shared-resource nature of our computing environment. As an indicative example, 
on the MUTAG dataset \method{} completed in $14.6 \pm 1.8$\,s per 
fold, compared to $5.7 \pm 1.9$\,s for GNN and $0.1$\,s for the 
graph kernel (the latter reported as total time divided by number 
of folds, hence no variance estimate). GNNs benefit from GPU 
acceleration, which explains their competitive training times 
despite their architectural complexity. The graph kernel's speed 
on MUTAG reflects both the small dataset size and its C++ 
implementation; on larger datasets kernel methods can become 
substantially slower due to their super-linear computational 
complexity with respect to graph size~\citep{Kriege2020}.

\section{Discussion and Conclusion}\label{sec:discussion}


\method{} demonstrates competitive performance on balanced graph classification tasks from the \tud{} collection, achieving the best accuracy among all evaluated methods on 6 out of 12 balanced datasets. The most substantial improvements occur on SYNTHETIC ($+13.7$ percentage points over the best competitor) and PROTEINS\_full ($+5.9$ points), while results on smaller or structurally simpler datasets are mixed.

The empirical evidence seems to suggest that \method{}'s performance advantage tends to grow with average graph size. We believe this behaviour is connected to the origins of the \method{} framework: the underlying regression algorithm \citep{meggio2024pathboost} was designed for the TMQM dataset of transition metal compounds, molecular graphs with diverse and often large structures. Path-based features are particularly well suited to such graphs because longer paths can capture richer structural patterns---ring structures, branching patterns, and extended functional groups, that are informative for property prediction. Moreover, the boosting procedure provides natural sparsification: even when larger graphs give rise to a combinatorially large number of candidate paths, \method{} selects only the most informative ones at each iteration, expanding the feature space incrementally rather than exhaustively. This greedy, residual-driven selection limits the effective model complexity, mitigating the risk of overfitting that would arise from naively enumerating all available paths.

We also believe that GNNs based on message-passing may be at a disadvantage on larger graphs due to a known depth dilemma: with few layers, the receptive field of each node is limited, potentially missing long-range structural patterns; yet increasing depth risks over-smoothing ~\citep{Li2018DeeperII}, where node representations converge and lose discriminative power. The standard GNN configurations used in the \tud{} benchmark (up to 5 layers) may therefore be insufficient for graphs with 100 or more nodes, as seen in the SYNTHETIC dataset. This complementary behavior suggests that the choice between \method{} and GNN approaches should be guided by dataset characteristics, particularly graph size and structural complexity, rather than assuming universal superiority of either approach.

A distinguishing feature of \method{} is the explainability it inherits from the underlying path-based boosting framework. At each boosting iteration, the algorithm explicitly identifies which labelled path drives the model update, and the contribution of each path to the overall prediction can be quantified through variable importance scores. This is in direct contrast to GNNs, where the learned representations entangle graph substructures within dense vector spaces, making it difficult to attribute predictions to specific structural features without post-hoc explanation methods.

In many applications, this explainability has practical value; for example, in the molecular study of \citet{meggio2024pathboost} the selected paths correspond to specific atomic sequences within molecules, allowing domain experts to evaluate whether the model has identified chemically meaningful patterns. The original regression framework demonstrated this capability on the TMQM dataset, where the most important paths corresponded to ring structures and metal-centre neighbourhoods that are known to influence molecular properties. While a detailed explainability analysis for classification tasks is beyond the scope of this work, the same path-level importance mechanism is available in \method{} and could provide similar insights in other future studies.

\section*{Declaration of competing interest}
The authors declare that they have no known competing financial 
interests or personal relationships that could have appeared to 
influence the work reported in this paper.

\section*{Data availability}
All datasets used in this study are publicly available through the 
TUDatasets collection \citep{TUD_paper}.

\section*{Code availability}
The source code for \method{} is available at  \url{https://github.com/Claudio-Me/Path_Based_Gradient_Boosting_for_Graph_Based_Prediction.git}.

\section*{Funding}
The authors acknowledge funding from the following sources: EU Horizon 2020, through the MSCA CompSci, project n. 945371 (CM); Research Council of Norway, through the Centre of Excellence Integreat, project n. 332645 (JP, RDB); Research Council of Norway, through the FRIPRO Plumbin’, project n. 323985 (RDB).

\bibliographystyle{plainnat}
\bibliography{refs}

\end{document}